%% file: main.tex
\documentclass[letterpaper, 10 pt, conference]{ieeeconf}  

\IEEEoverridecommandlockouts                              

\input{config}

\input{math_commands}

\title{\modelname: Perception Alignment for Human-Robot Collaboration}
\input{authors}

\begin{document}

\maketitle
\thispagestyle{empty}
\pagestyle{empty}

\begin{abstract}
Recently, \acp{llm} have shown strong potential in facilitating human-robotic interaction and collaboration. However, existing \ac{llm}-based systems often overlook the misalignment between human and robot perceptions, which hinders their effective communication and real-world robot deployment. To address this issue, we introduce \modelname, a unified system designed to achieve both perceptual alignment and human-robot collaboration. At its core, \modelname employs \ac{sg3d} as its explicit and innate representation. This enables the system to leverage \ac{llm} to break down complex tasks and allocate appropriate tools in intermediate steps to extract relevant information from the \ac{sg3d}, modify its structure, or generate responses. Importantly, \modelname incorporates an automatic mechanism that enables perceptual misalignment correction with users by updating its \ac{sg3d} with online interaction. 
\modelname achieves comparable performance with the data-driven models in ScanQA in a zero-shot manner. Through comprehensive experiments across 10 real-world scenes, \modelname demonstrates its effectiveness in establishing common ground with humans, realizing a success rate of 61.9\% in alignment tasks. It also significantly improves the success rate from 3.7\% to 45.68\% on novel tasks by transferring the knowledge acquired during alignment.
\end{abstract}

\section{Introduction}\label{sec:intro}
\begin{figure}[t!]
  \centering
  \includegraphics[width=\linewidth]{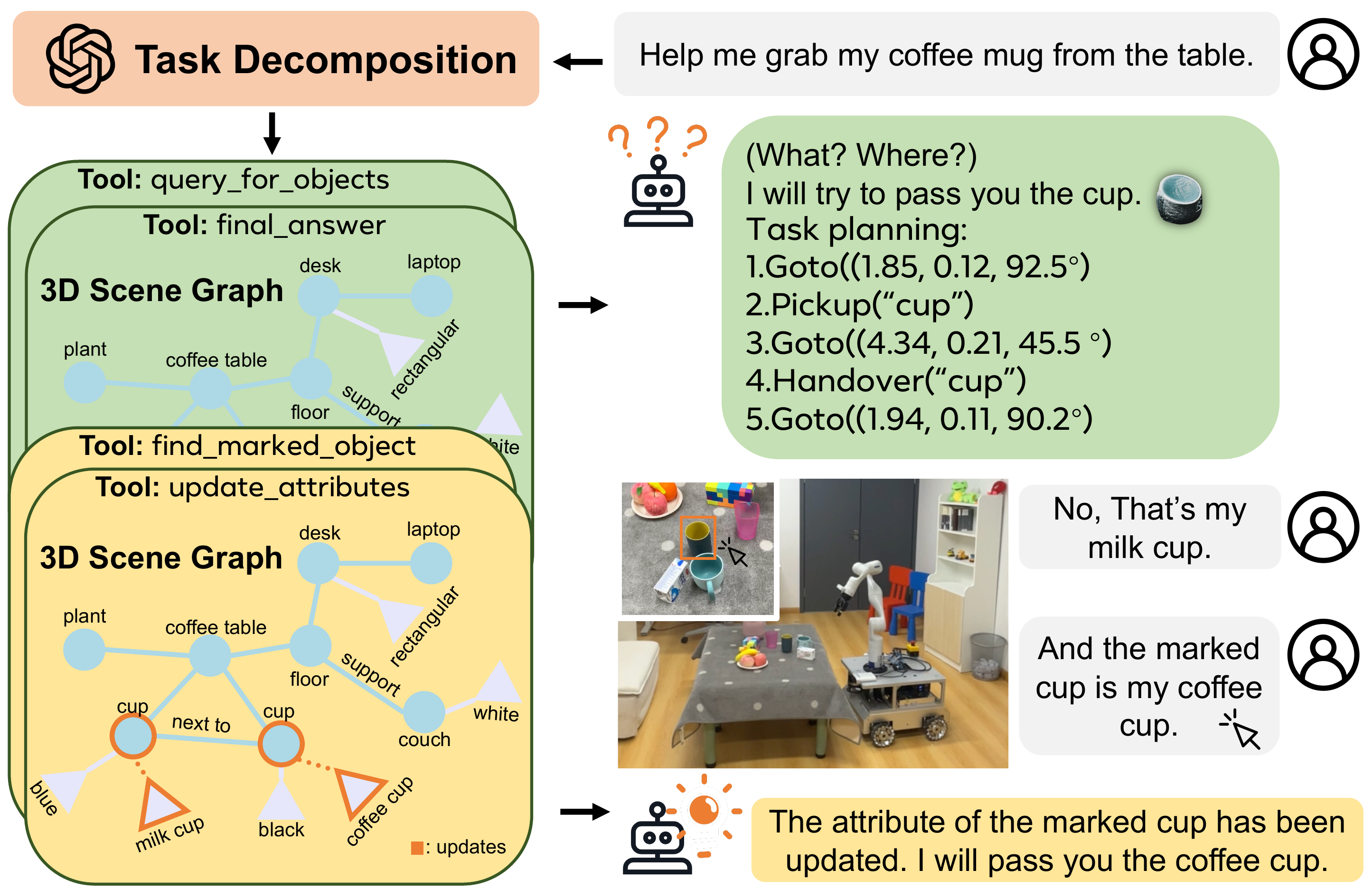}
  \caption{\textbf{Overview of \modelname.} Leveraging \acf{sg3d} as its representation, \modelname decomposes complex tasks with \ac{llm}s and takes actions with our designed tools in intermediate steps. It interacts with humans through natural language and non-verbal mouse clicking to enhance object references, capable of facilitating human-robot collaboration and perceptual alignment by automatically modifying the data stored in \ac{sg3d}.}
  \label{fig:teaser}
\end{figure}

Imagine a household robot assisting humans in their homes, receiving an instruction \textit{``help me grab my coffee mug from the kitchen''}, as depicted in \cref{fig:teaser}. While this task seems routine, complications arise if the robot lacks knowledge of where the kitchen is or, more importantly, cannot discern the coffee mug among the various containers in the 3D environment. Achieving an accurate understanding of such scenes, including their semantic attributes, spatial arrangements, and personalized references like ``one's coffee mug,'' is essential for the robot to effectively employ its planning, navigation, and manipulation skills. Effective communication between humans and robots, whether simple or complex, relies on establishing a common ground to facilitate human-robot collaborations, with perception serving as its foundational milestone. Despite the recent popularity of applying \ac{llm} in robotics~\cite{silver2022pddl,song2023llm,brohan2023can,ding2023integrating,liu2023reflect}, they overlook this fundamental aspect.

To achieve effective human-robot collaboration, a robotic system must meet several key requirements: (1) the ability to construct a compact representation of the 3D environment; (2) the versatility to handle various tasks under natural-language instructions in a zero-shot manner; and (3) the flexibility to quickly align with users through natural interactions, all within a coherent system.

However, existing systems fall short of these requirements: they are not yet capable of perfectly perceiving real-world scenes and are not customized for individual users, making it difficult to learn personalized concepts such as naming conventions or preferences shaped by diverse cultures and lifestyles. Thus it is essential for the robot to quickly correct perceptual misalignment with users to enhance human-machine cooperation in completing real-world tasks.

In this paper, we propose \modelname, a system designed to meet the aforementioned criteria. Our framework takes a collection of posed images as input, reconstructs the 3D scene and constructs a \acf{sg3d}~\cite{armeni20193d,chen2019holistic++, wald2020learning,rosinol2021kimera} as its data structure. The \ac{sg3d} encapsulates hierarchical topology and key information necessary for 3D reasoning, including object categories, attributes, states, and spatial relationships in the scene. 

Taking advantage of the explicit and explainable advantage of \ac{sg3d}, we leverage \acs{llm} to break down complex tasks---such as question answering, task planning, and captioning---into intermediate steps, and allocate appropriate tools for completing these steps. The toolset is designed to extract
relevant knowledge from the 3DSG, make modifications, or generate responses based on the user input, supporting a wide range of open-world reasoning tasks. Additionally, we have incorporated an alignment mechanism that autonomously recognizes users' intention of alignment and triggers a process to update the \ac{sg3d} on the fly. For higher efficiency in human-robot interaction, we have developed a user-friendly graphical interface that allows users to freely interact with the scene by dragging, zooming-in/out, changing views, marking objects and asking free-form questions at will.

We conduct extensive experiments to demonstrate \modelname's capabilities in human-robot collaboration and alignment. Results show that, even in a zero-shot manner, it achieves comparable performance with data-driven methods on ScanQA~\cite{azuma2022scanqa}. More importantly, we perform real-world alignment experiments of varying difficulties between \modelname and humans to systematically assess its ability to establish common ground with humans. Our model realizes a success rate of 61.9\% for the alignment tasks while providing a smooth interaction experience (64.87\% per-step user satisfaction rate). It further demonstrates the capability to transfer the acquired knowledge to novel tasks by improving their success rate from 3.70\% to 46.58\%.

\section{Related Work}
\label{sec:related_work}

\textbf{Human-Robot Alignment}\quad
Significant attention has recently been focused on the Human-machine alignment~\cite{hoc2000human,christian2021alignment}, especially for aligning \acs{llm}s with human intentions and values~\cite{ji2023ai,casper2023open} with \ac{rlhf}~\cite{christiano2017deep,ouyang2022training,rafailov2024direct} or \ac{sft}~\cite{ji2024aligner}.
In robotics, the notion of human-robot alignment centers on improving their coordination in real-world scenarios. Previous efforts have been devoted to enhancing effective human-robot interaction through dialogue, including generating help requests~\cite{tellex2014asking}, seeking oracle in planning~\cite{nguyen2019help,thomason2020vision,padmakumar2022teach}, following embodied instructions~\cite{gao2022dialfred}, and resolving the uncertainty of  LLM-based planners~\cite{ren2023robots}. Despite their progress in establishing natural language communications with humans~\cite{yao2022webshop,brohan2023can,deng2024mind2web}, one limitation lies in the presumption that the robot and humans have reached a common ground, overlooking the fact that the robotic perception capabilities remain far from perfect to date. In this paper, we propose a systematic framework that enables the robot to align with humans both preemptively and during collaborative tasks. The alignment process is facilitated through natural interactions, \ie, natural language or a virtual interface, thereby endowing robots with correct perception aligned with individual human perspectives.

\textbf{\acs{llm} in Robotics}\quad
Previous research has effectively leveraged pre-trained \acs{llm}s’ in-context learning abilities for embodied agents to generate actionable task plans~\cite{silver2022pddl,song2023llm,brohan2023can,zeng2022socratic,liu2023llm,huang2023inner}, recover from failure~\cite{ding2023integrating,liu2023reflect,wang2023voyager,wang2024describe,zhi2024closed}, perform low-level control~\cite{liang2023code}, or specify reward functions~\cite{tam2022semantic,kwon2023reward,du2023guiding,hu2023language}. In order to enable these language models to perceive physical environments, visual information is either decoded with grounded models~\cite{huang2023grounded} or directly treated as input by \ac{mmlm}~\cite{driess2023palm,li2023blip2,openai2023gpt}. However, the power of these foundation models is often limited in separate stages of training or inference, leaving the potential for humans to \textit{teach} the robots unexplored. Thus our framework is proposed to facilitate the robots' potential to evolve in understanding the 3D world through interactions with humans. We utilize the structured representation of \ac{sg3d}~\cite{armeni20193d,chen2019holistic++, wald2020learning,rosinol2021kimera} and its compatibility with \ac{llm} to boost human-robot coordination, a critical aspect for deploying personalized humanoids in real-world scenarios.

\textbf{\acs{llm} in 3D Scene Understanding}\quad
The popularity of \acs{llm}s has recently spurred the development of 3D scene understanding in various tasks, \eg, object referral~\cite{chen2020scanrefer,achlioptas2020referit3d}, captioning~\cite{chen2021scan2cap,yuan2022x}, vision-language-navigation~\cite{ma2019self,hong2021vln} and reasoning~\cite{azuma2022scanqa,ma2022sqa3d}. The signature efforts like 3D-LLM~\cite{hong20233d}, Chat-3D~\cite{wang2023chat}, LEO~\cite{huang2023embodied} and 3DMIT~\cite{li20243dmit} investigate alternatives to incorporate the multi-modal inputs, \eg, 3D point clouds, images, and texts, into a pre-trained \ac{llm} and further fine-tune the model on more data for downstream tasks. In this paper, we devise a framework capable of performing 3D reasoning tasks important for human-robot collaboration, \ie, embodied question answering, task planning, and captioning, all in a zero-shot manner. The agent utilizes \ac{sg3d} as explicit representations and harnesses the extensive reasoning capabilities of foundation models to interact with humans and accomplish comprehensive tasks.

\section{Method}
In this section, we introduce \modelname, which aims to perceive the 3D world, tackle the 3D reasoning tasks in a zero-shot manner, and align with human perception through natural interactions. We first present the reconstruction of 3D scenes and the construction of \ac{sg3d} in \cref{sec:method_scene}. We then illustrate the system design in \cref{sec:method_agent} with available tools in \cref{tab:tools}.

\begin{figure*}[t]
     \centering
     \includegraphics[width=0.95\textwidth]{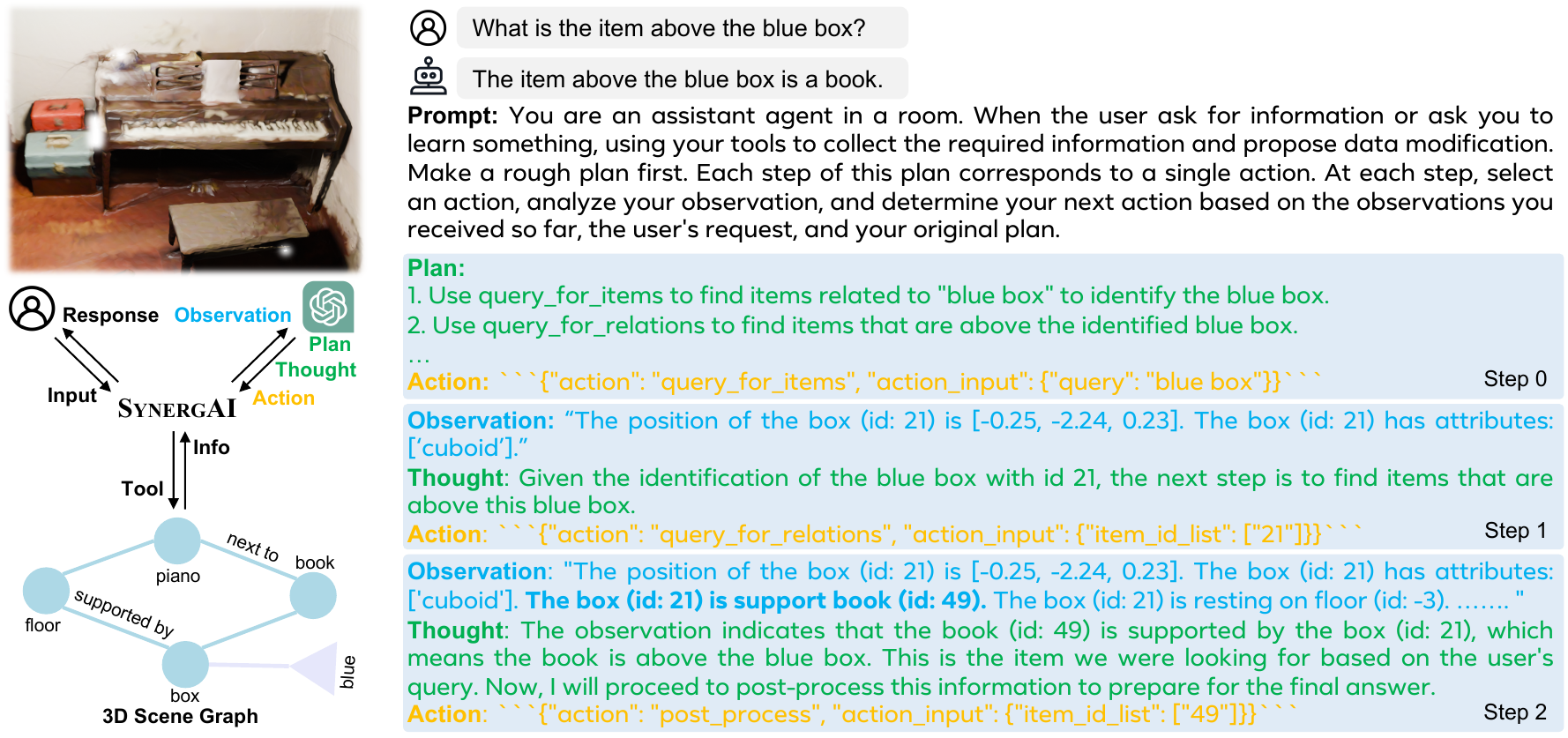}
     \caption{\textbf{The design of \modelname and an example interaction}. \modelname represents 3D scene with \ac{sg3d}s and leverages \ac{llm}s to respond to user inputs. It is first prompted to generate a \emph{plan}, which effectively decomposes the input task into sub-tasks to be solved in a sequential process. At each step, \modelname selects a \emph{tool} as its \emph{action} based on the \emph{observation}, which contains the results of the previous actions. In this example, the system identifies the correct object of relationship \textit{``on the blue box''}, but incorrectly recognizes it as a book, where \textit{perception misalignment} happens.}
     \label{fig:agent_design}
\end{figure*}

\subsection{3D Scene Reconstruction and \texorpdfstring{\acf{sg3d}}{}}
\label{sec:method_scene}
From posed RGBD images,  the 3D mesh of a scene can be reconstructed through either depth fusion and marching cubes algorithm~\cite{lorensen1987marching} following ScanNet~\cite{dai2017scannet}, or via neural rendering with \sota methods like MonoSDF~\cite{yu2022monosdf}. Subsequently, we obtain the object instances and their semantic labels by employing 3D instance segmentation~\cite{zhu2023vista}, 2D image classification~\cite{li2023blip2} and multi-view association~\cite{gu2023conceptgraphs}.

\ac{sg3d} have recently emerged as an effective world representation for robotics~\cite{rosinol2021kimera,kurenkov2021semantic,hughes2022hydra,agia2022taskography,ravichandran2022hierarchical}, capturing a hierarchically organized semantic graph representation of an environment with the versatility to encode the object states and spatial relationships. The data structure is suitable for parsing and we convert the graph to the \acs{llm} inputs similar to JSON serialization (see example observations in \cref{fig:agent_design}). The \ac{sg3d} is defined as a hierarchical graph $\mathcal{G}=(\mathcal{V},\mathcal{E})$, where each node $v\in \mathcal{V}$ represents one distinct 3D object instance, denoted by its position, size, and attributes. The edges $\mathcal{E}$ represent spatial relationships between nodes.

To construct the scene graph $\mathcal{G}$, we first instantiate the nodes with the instance predictions from the reconstructed mesh and assign object classes with their corresponding semantic labels. To acquire the semantic-rich object states and attributes, we utilize  3D object segmentation to identify its occurrence in the multi-view images through rendering. The images are then cropped with the rendered bounding boxes and processed through BLIP2~\cite{li2023blip2} to generate information about the object's color, texture, shape, material and affordance. Following prior work~\cite{achlioptas2020referit3d,wald2020learning,jia2024sceneverse}, we utilize the positional and size information of objects within the scene to model the following spatial relations:
\begin{itemize}
\item \textbf{Horizontal proximity} reflects the distance of horizontal object placements, \eg, \texttt{close} or \texttt{far}.
\item \textbf{Vertical proximity} encompasses both in-contact relationships (\eg, \texttt{support}, \texttt{inside}, \texttt{embed}), and non-contact ones (\eg, \texttt{above}, \texttt{below}). 
\item \textbf{Allocentric} relationships describe the directional relations like \texttt{left, right, in front of}, \etc, which are contextually dependent on the robots' viewing direction. Our framework \textit{dynamically} updates the allocentric relationships based on its current position and viewing angle.
\end{itemize}

We traverse all the object nodes to calculate spatial relationships, which undergo an automatic verification procedure to rectify incorrect ones.

\begin{table*}[!t]
    \scriptsize
    \centering
    \caption{\textbf{The toolset for \modelname}. They are designed as Python APIs, with the top five tools for 3D reasoning, the following four for alignment, and the last two for generating responses to the user.}
    \label{tab:tools}
    \resizebox{\linewidth}{!}{
    \begin{tabular}{llll}
        \toprule
        \textbf{Tool} & \textbf{Input} & \textbf{Return} & \textbf{Description}\\
        \midrule
        \texttt{query\_for\_objects} &String, $\mathcal{G}$ &List[Object] &Collect the objects mentioned in a user input.\\
        \texttt{query\_for\_relations} &List[Object], $\mathcal{G}$ &List[Relation] &Collect the relations associated with a list of objects.\\
        \texttt{find\_marked\_object} &Click, $\mathcal{G}$ &Object &Collect the information of the object marked by the user.\\
        \texttt{calculate\_mid\_point} &List[Point], $\mathcal{G}$ &Point &Calculate the midpoint of a list of Points.\\ 
        \texttt{find\_object\_closest} & Point, $\mathcal{G}$ &Object&Collect the object closest to a point.\\ \midrule
        \texttt{update\_name} &List[String], List[Object], $\mathcal{G}$ &$\mathcal{G}$, List[Object]&Update the labels of a list of objects. \\
        \texttt{update\_attributes} &Object, List[String], $\mathcal{G}$ &$\mathcal{G}$, List[Relation]& Update the attributes of an object. \\
        \texttt{add\_relation} &Object, Object, Relation, $\mathcal{G}$ &$\mathcal{G}$, List[Relation]& Add a relation between two objects.\\
        \texttt{delete\_relation} &Object, Object, Relation, $\mathcal{G}$ &$\mathcal{G}$, List[Relation] &Remove a relation between two objects. \\  \midrule
        \texttt{post\_process} &List[Objects] &List[Objects] & Return the relevant information for the \ac{gui}. \\
        \texttt{final\_answer} &String & String & Return the final response for the input. \\
        \bottomrule
    \end{tabular}
    }
\end{table*}

\subsection{System Design}
\label{sec:method_agent}
Given the \ac{sg3d} as the scene representation, we aim to develop a robot system that can communicate with humans, perform 3D reasoning tasks, and align with human perception by leveraging the power of \ac{llm}. However, this presents two major challenges: 1) The intricate nature of the tasks and the 3D scenes makes it difficult to directly utilize the full \ac{sg3d} for complex reasoning, even for \ac{llm}. 2) Language-based interaction alone is insufficient for efficiently referencing objects in the presence of erroneous labels and relations, which is essential to accomplish the alignment tasks. The following explains how the design of \modelname addresses these two challenges.

\textbf{Task Decomposition}\quad
We begin by noting that it's more efficient to tackle complex tasks step-by-step, similar to \ac{cot}~\cite{wei2022chain}, and most intermediate reasoning steps exhibit locality–they are solvable once the relevant information is retrieved from the 3D scenes. Therefore, our idea is to prompt the system to decompose the complex tasks into intermediate steps and progressively gather relevant sub-graphs from the \ac{sg3d} to tackle them. As illustrated in \cref{fig:agent_design}, upon receiving user input, \modelname invokes a sequential process with \ac{llm}, where at each step the system receives an \emph{observation}, generates a \emph{thought} and selects an \emph{action}. The actions are calls to a set of APIs called \emph{tools}, and the \emph{thought} is the reasoning process and rationale behind choosing the next action. The \emph{observation} contains the summarized information in the previous action, \eg, the sub-graph retrieved from \ac{sg3d}. This process continues unless the agent gathers enough information and selects the termination tool \texttt{final\_answer}, which finishes the reasoning process and returns the final response to the user. 

Note that the agent is prompted to compose a \emph{plan} that outlines the task decomposition at the first step in \cref{fig:agent_design}. By composing such a plan, the agent is guided to recognize the user's intent, thereby reducing its workload in later steps. Moreover, by conditioning on this plan, the agent can proceed in a top-down fashion and select actions accordingly. Note that the actual executed actions may deviate from the original plan: the agent is capable of modifying its plan during thoughts when new observation reveals mistakes. Such flexibility turns out to be essential for complex 3D reasoning tasks where they can be resolved in different ways.

\textbf{Observation}\quad Starting from the second step, \modelname generates an observation that summarizes the information retrieved from the \ac{sg3d} by the last action or indicates errors occurred in calling tools. The agent utilizes these observations to generate thoughts and decide its next action. To transfer the retrieved sub-graph from \ac{sg3d} into inputs compatible with \ac{llm}s, we render retrieved objects and the associated relations into sentences using templates. For example, we use ``\texttt{The \{object.name\} (id: \{object.id\}) has attributes: \{object.attributes\}.}" as the template for rendering the attribute-related information of the user-specified object.

\textbf{Human-Robot Interaction}\quad As previously mentioned, object references are essential for achieving alignment as they provide a common ground for communication. However, under erroneous perception, users may struggle to reference objects with incorrect labels and attributes in 3D environments via pure language-based interactions. Motivated by the higher efficiency of non-verbal cues in object reference than pure language in human communication~\cite{mcneill2012language,arbib2008primate,chen2021yourefit}, we address this challenge by implementing a \ac{gui}, which includes the reconstructed 3D scene, the segmentation, and the \acl{sg3d}. It further allows users to mark objects by clicking in the 3D scene or the nodes in the \ac{sg3d}. This non-verbal interaction is robust against semantic errors and can function if an object is properly segmented. Users can thus refer to an item as ``the marked object" in their inputs after clicking on it, which significantly reduces their workload in human-robot alignment.

\textbf{Tools} \quad
The final piece of \modelname is the tools available for each intermediate step, which are a set of Python APIs summarized in \cref{tab:tools}. The tools are designed to extract relevant information from the \ac{sg3d}, modify its structure, or generate responses based on the user input. Our designed tools can be categorized into three purposes, 3D reasoning, alignment, and response generation, but during the sequential task-solving process, \modelname automatically identifies the user intent and selects tools accordingly based on the user inputs. This means \modelname can decide if it is required to perform 3D reasoning or align with humans based on the users' natural language instructions, without the need for mode shifting or explicit task specifications.


\textbf{Implementation}\quad We develop \modelname based on the LangChain~\cite{langchain2022} framework. More specifically, we rely on \texttt{GPT-4 turbo} as the underlying \ac{llm} backend, and LangChain manages the sequential process, including generating step-wise prompts, parsing the agent's output, and executing the tools. The step-wise prompt combines a template with the latest observation, doc-strings of tools, and all historical observations, actions, and thoughts.

\section{Experiments}

\label{sec:exp}
\subsection{Human-Robot Collaboration in Zero-shot 3D Reasoning}

\begin{figure}
    \centering
    \includegraphics[width=\linewidth]{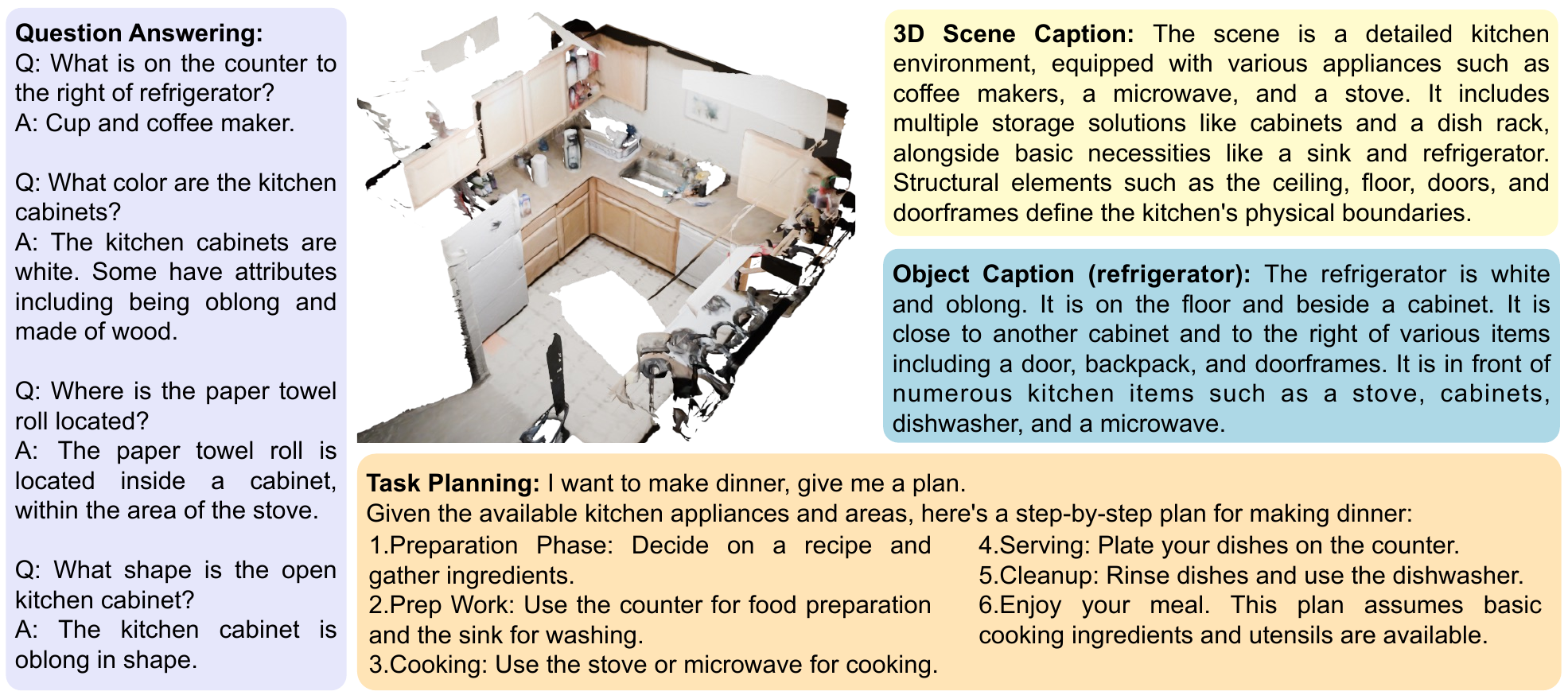}
    \caption{\textbf{Qualitative results of 3D Reasoning Tasks.}}
    \label{fig:qual}
    \vspace{-1em}
\end{figure}

\textbf{Dataset and Metrics}\quad To extensively evaluate \modelname's capability in high-level human-robot collaboration with language-guided interactions, we evaluate its zero-shot 3D reasoning task performance, with qualitative demonstrations in object captioning, scene captioning, question-answering and task planning in \cref{fig:qual}. To quantitatively evaluate its performance, we select the one that best reflects its reasoning capability, \ac{qa} task. We utilize the ScanQA~\cite{azuma2022scanqa} benchmark to test our system. Since we evaluate our system in a zero-shot manner \textit{without alignment}, we directly test its performance on its validation set, encompassing a total of 4675 scene QA tasks from 71 3D scenes in ScanNet~\cite{dai2017scannet}. Following common practice~\cite{azuma2022scanqa}, we assess the \ac{qa} performance using Exact Match (EM), CIDEr, BLEU-1, METEOR, and ROUGE.

\textbf{Performance Analysis}\quad From \cref{fig:qual}, we can see \modelname is capable of accomplishing various 3D reasoning tasks in a zero-shot manner. \cref{tab:scanqa} demonstrate the quantitative results on ScanQA, where our model achieves comparable performance to the methods \textit{fine-tuned} on ScanQA, yet fails short on the metric EM. The discrepancy can be attributed to the fact that our system generates answers by leveraging the power of \ac{llm}, which aligns more closely with human preference. For instance, given the question ``\textit{What is sitting on top of the toilet tank lid?}'', our response is ``\textit{A towel is sitting on top of the toilet tank lid.}'', whereas the ground-truth response is simply ``\textit{towel}''. The differentiation in format significantly affects metrics like EM. 

\begin{table}[t!]
    \centering
        \caption{\textbf{Zero-shot Performance on ScanQA.} Our model reaches comparable performance with fine-tuned baselines.}
        \resizebox{\linewidth}{!}{
        \begin{tabular}{lccccccc}
            \toprule
            Model&EM&CIDEr&BLUE-1&METEOR&ROUGE\\
            \midrule
            \textbf{\textit{Fine-tuned}} \\
            VoteNet+MCAN & 17.3 & 54.7 & 28.0 & 11.4 & 29.8 \\
            ScanRefer+MCAN & 18.6 & 55.4 & 26.9 & 11.5 & 30.0 \\
            ScanQA & 21.0 & 64.9 & 30.2 & 13.1 & 33.3 \\
            Flamingo (MultiView) & 18.8 & 55.0 & 25.6 & 11.3 & 31.1 \\
            BLIP2 (MultiVIew) & 13.6 & 45.7 & 29.7 & 11.3 & 26.6 \\
            3D-LLM (Flamingo) & 20.4 & 59.2 & 30.3 & 12.2 & 32.3 \\
            \midrule
            \textbf{\textit{Zero-shot}} \\
            Ours & 11.4 & 57.9 & 30.1 & 12.9 & 30.4 \\
            \bottomrule
        \end{tabular}
        }
        \label{tab:scanqa}
        \vspace{-1em}
        \end{table}

\subsection{Human-Robot Alignment}
\textbf{Setup}\quad We systematically assess \modelname's capability in achieving perceptual alignment with humans spanning 10 real-world scenes sourced from the ScanNet~\cite{dai2017scannet} dataset. The tests include two phases, \ie, \textbf{alignment tasks} and \textbf{knowledge transfer}. In the first phase, we devise 42 alignment tasks targeted at the perception errors related to object naming, shape, material, and spatial relations. The tasks are designed in the form of question-answering and the objective is to correct the system's perception through interacting with it in the \ac{gui}, ensuring that the system's final responses ultimately align with humans. The tasks are categorized into the \emph{EASY} (25 tasks) and \emph{HARD} (17 tasks) splits, depending on the number of items involved in the alignment tasks. In the second phase, we design 27 novel tasks dependent on the misaligned concepts to measure if the knowledge acquired from the alignment can be transferred.

We engage the participation of 10 human subjects for the alignment experiment. They first undergo a preparatory session to become acquainted with our system under instructions, following which each participant is assigned tasks across 3 scenes. During the alignment experiments, subjects are tasked to inspect and correct the perception of the agent within the \ac{gui} using both natural language and mouse-click interactions. Participants are required to provide a binary rating for the system's response to each interaction and assess the success of the alignment after each task.

\textbf{Metrics}\quad We evaluate the success rate of the alignment task with both human (\textit{\srh}) and \ac{llm} (\textit{\srl}) judgments. The answer accuracy of our system before the alignment is evaluated by \ac{llm} as a baseline, denoted as \textit{\sri}. The ratio of reasonable responses (\textit{\rrr}) from our system during the interactions is judged by the human subjects. For task difficulty, we report the average number of interactions required to complete per task (\textit{\#Inter./Task}) and the average number of actions that the system executes to generate responses in one interaction (\textit{\#Action/Inter.}). The query ratio (\textit{QR}) is reported as the ratio of tokens required to encode the extracted relevant sub-graphs and the full \ac{sg3d}. The knowledge transfer tasks are also measured by the success rate of question-answering. We provide the mean and 95\% confidence intervals for all the metrics.

\begin{figure}[t]
    \centering
    \includegraphics[width=\linewidth]{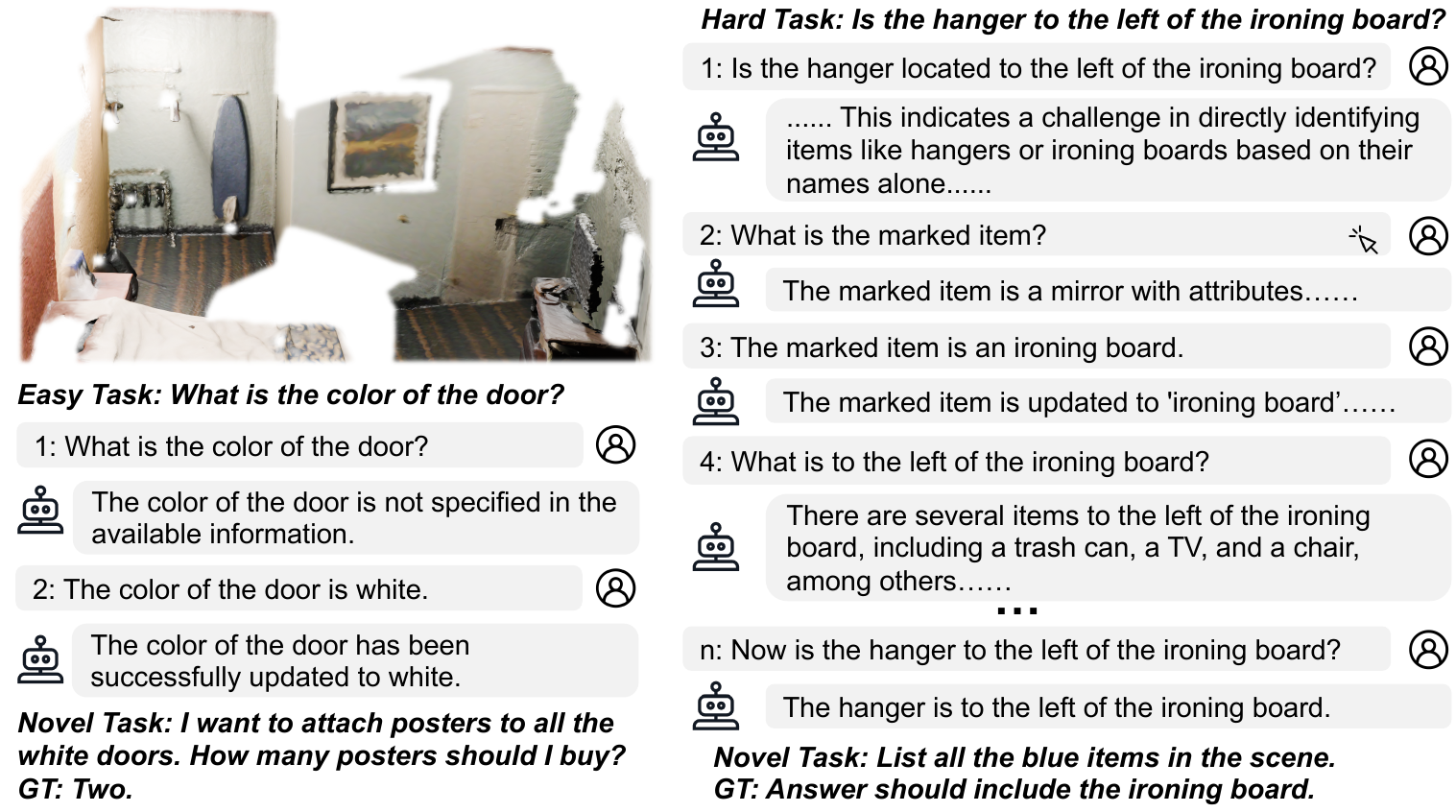}
    \caption{\textbf{Examples of human-robot alignment}. Humans solve the \emph{EASY} task within shorter interaction steps compared with the \emph{HARD} task, where the user checks and corrects the label of the ironing board by clicking (the 2\textsuperscript{nd} \& 3\textsuperscript{rd} user inputs). \emph{Novel} tasks are designed such that knowledge from the alignment is required for their completion.} 
    \label{fig:task_example}
    \vspace{-1em}
\end{figure}

\begin{table*}[!t]
    \centering
    \scriptsize
    \caption{\textbf{Quantitative results of human-robot alignment.} ``SR'' denotes the success rate for the alignment tasks, ``RR'' for the rate of reasonable responses and ``QR'' for the query ratio of the \ac{sg3d}.}
    \label{tab:alignment_results}
    \resizebox{0.9\linewidth}{!}{
    \begin{tabular}{ccccccccc}
        \toprule
        & \multicolumn{4}{c}{Alignment Task} & \multicolumn{3}{c}{Task Difficulty}\\
        \cmidrule(lr){2-5}\cmidrule(lr){6-8}
         & \sri (\%) & \srh (\%) & \srl (\%) & \rrr (\%) & \#Inter./Task & \#Action/Inter. & QR (\%)\\
        \midrule
        EASY & 8.00 & 91.18±9.89 & 72.00±14.14 & 74.06±9.19 & 3.38±0.68 & 3.23±0.17 & 2.55±0.78\\
        HARD & 0.0 & 72.48±14.2 & 47.06±20.15 & 51.36±9.49 & 6.45±1.07 & 3.30±0.24 & 3.86±1.23\\
        \midrule
        OVERALL & 4.76 & 83.61±8.36 & 61.90±11.83 & 64.87±7.31 & 4.65±0.74 & 3.26±0.13 & 3.08±0.68\\
        \bottomrule
    \end{tabular}
    }
\end{table*}

\begin{figure*}
 \centering
 \resizebox{0.95\linewidth}{!}{
 \begin{subfigure}[b]{0.3\textwidth}
     \centering
     \includegraphics[width=\textwidth]{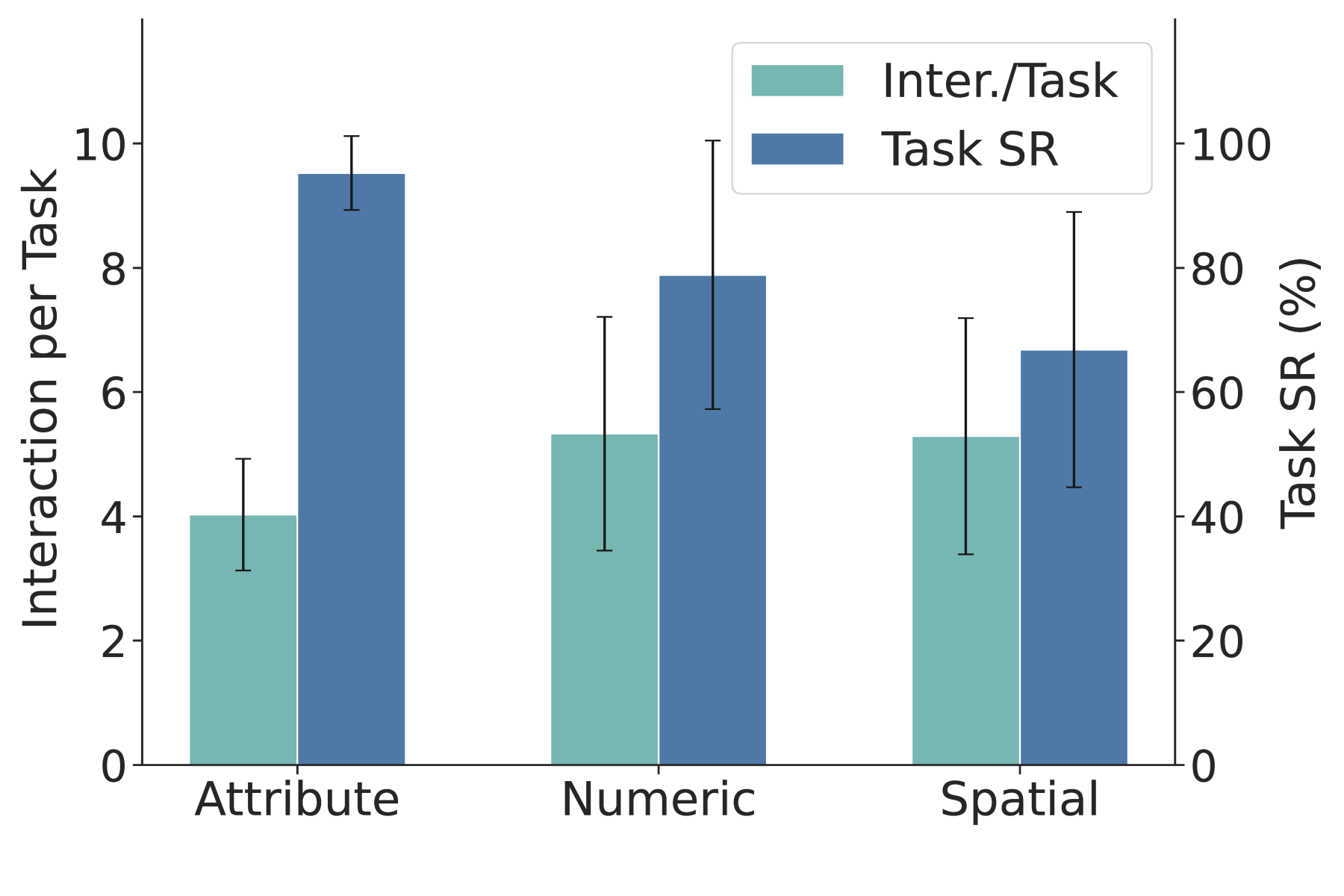}
     \caption{Success rate and \#Interations by task.}
     \label{fig:sr_category}
 \end{subfigure}
 \hfill
 \begin{subfigure}[b]{0.3\textwidth}
     \centering
     \includegraphics[width=\textwidth]{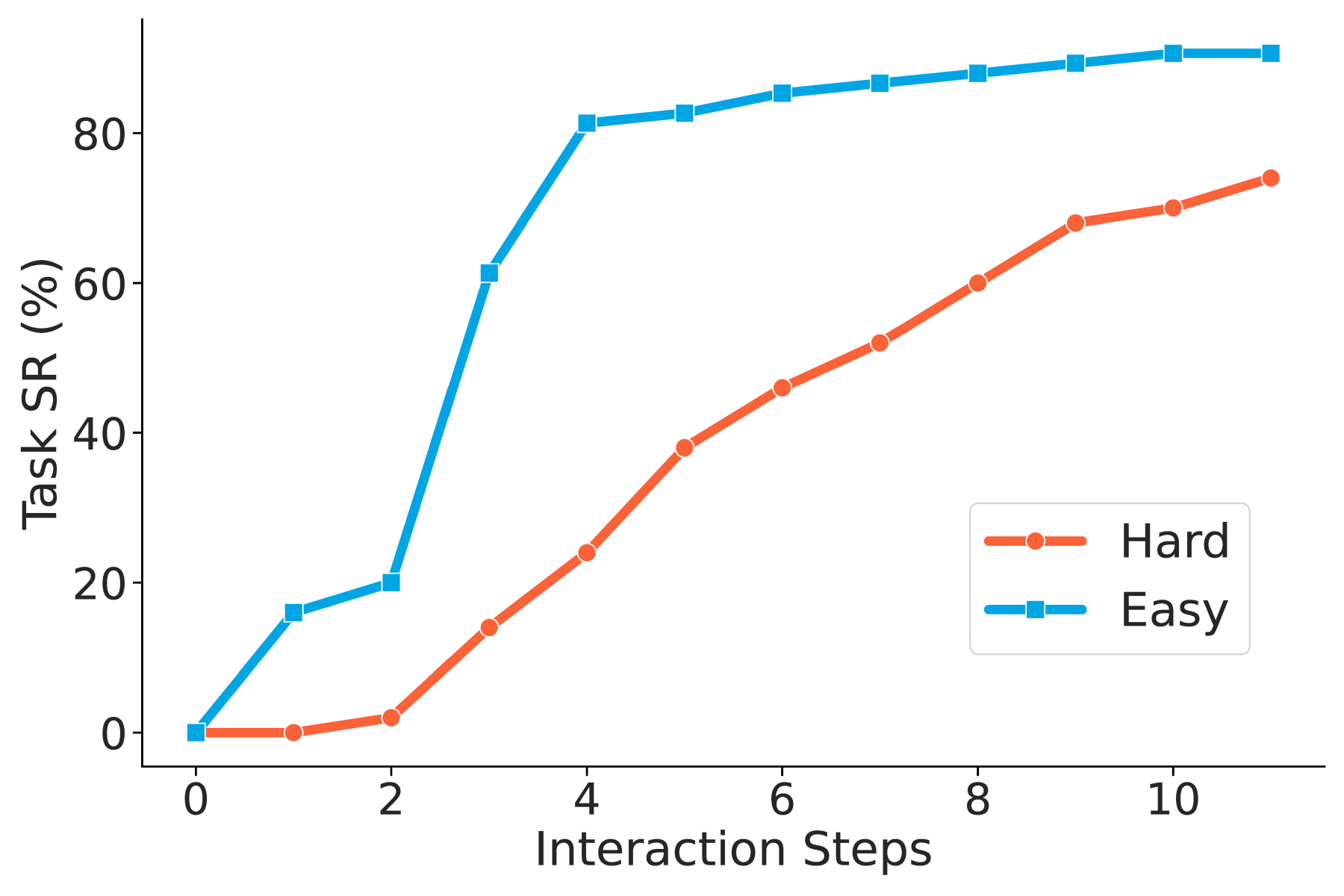}
     \caption{Task success rate vs. \#interactions.}
     \label{fig:sr_step}
 \end{subfigure}
 \hfill
 \begin{subfigure}[b]{0.3\textwidth}
     \centering
     \includegraphics[width=\textwidth]{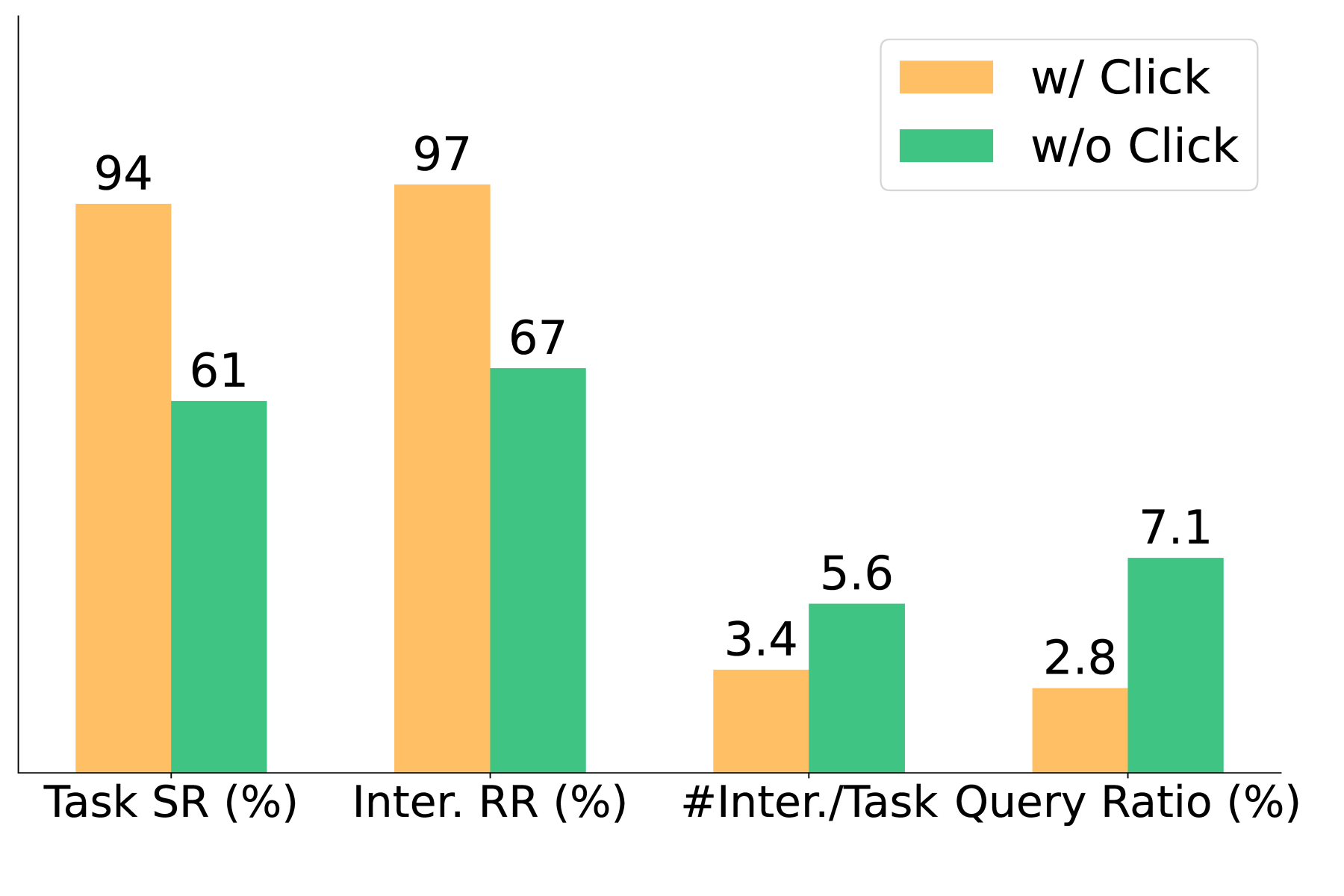}
     \caption{Ablation on mouse clicking.}
     \label{fig:gui_abl}
 \end{subfigure}
 }
\caption{\textbf{Statistics of alignment experiments.} (a) The success rate decreases for more complex tasks with increasing interaction steps required to achieve alignment. (b) The trend of task success rate as the interaction step increases. (c) The user interface impacts users' ability to reference objects and hinders the alignment performance when mouse clicks are not used. }
\vspace{-1em}
\label{fig:alignment_stats}
\end{figure*}

\begin{table}[ht]
    \centering
    \scriptsize
    \caption{\textbf{Knowledge Transfer to Novel Tasks.} Results are reported in Success Rate (\%) as measured by \ac{llm}.}
    \label{tab:transfer_exp}
    \begin{tabular}{cccc}
        \toprule
         & Init. & In-Context Prompting~\cite{brown2020language} & Ours \\
        \midrule
        EASY & 2.56 & 23.08 & 43.59 \\
        HARD & 4.76 & 28.57 & 47.62 \\
        \midrule
        OVERALL & 3.70 & 25.93 & 45.68 \\
        \bottomrule
    \end{tabular}
    \vspace{-1em}
\end{table}

\textbf{Results}\quad \cref{tab:alignment_results} and \cref{fig:alignment_stats} present the major results for the alignment experiments, and \cref{tab:transfer_exp} demonstrates the quantitative results for the knowledge transfer experiments.  Below we summarize the key observations.

\begin{itemize}
\item \textbf{Our system is capable of achieving perceptual alignment with humans}, which is validated by the task success rate from both human (\textit{\srh}) and machine (\textit{\srl}) evaluations. Meanwhile, it provides reasonable responses at each step of the interactions from the interaction satisfactory ratio.
\item \textbf{The task difficulty significantly impacts both success rates and efforts for alignment.} The \textit{\srh}, \textit{\srl} and \textit{\rrr} all present a notable gap between the \emph{EASY} and \emph{HARD} tasks. Efforts for alignment (\textit{\#Inter./Task}) are twice for the \emph{HARD} tasks, while the \textit{\#Action/Inter.} remains similar. This suggests a human tendency to decompose complex alignment to simpler tasks that impose a constant load for \ac{llm}-based agent. The \textit{Query Ratio} is also higher for \emph{HARD} tasks as they involve more objects in the \ac{sg3d}. \cref{fig:sr_category} shows an interaction step increase from tasks related to unary semantics, \ie, object attributes, to those involving n-ary information such as numeric counting or spatial relations, with success ratios consistently decreasing.
\item \textbf{The object reference plays a pivotal role in perception alignment.} We conduct an ablation study to verify the significance of mouse-clicking interactions. Results in \cref{fig:gui_abl} show a notable decline in the task success rate and an increase in the steps required for alignment without mouse clicking. This is attributed to the extra steps needed to reference the relevant objects, constituting a common ground for communication. Such findings align with our motivation for designing the \ac{gui} to facilitate efficient object marking, and underscore the necessity to develop improved human-robot interaction interfaces for future VR/AR and robot applications.
\item \textbf{Our model can transfer the knowledge acquired in the alignment to novel tasks}. We design a baseline model that leverages in-context prompting~\cite{brown2020language}, which takes the \textbf{ground-truth} knowledge for the alignment tasks as additional inputs in the prompt, to answer the questions in the novel tasks. On the contrary, our model directly takes the updated \ac{sg3d} from the alignment phase, though not perfect. As shown in \cref{tab:transfer_exp}, our improved results prove the effectiveness of explicitly updating the \ac{sg3d} during alignment, which leads to better systematic generalizability for novel tasks.

\end{itemize}

\subsection{Discussion and Limitation}
In this paper, we focus on evaluating the applicability of \acs{llm}-based agents to function \textbf{in real-world settings}, emphasizing their robustness to operate effectively under imperfect perceptions rather than showcasing their strengths \textbf{in ideal settings}. We believe this is fundamental for the practical use of \acs{llm}-based agents in the real world.

We identify the following limitations of our system. First, despite the remarkable reasoning capabilities of \acp{llm}, their tendency towards hallucination may lead to plans and responses that deviate from the \ac{sg3d}. Second, we design tools for our system to ensure their functionality completeness for existing tasks. This indicates our framework may not be able to perform certain ``novel'' operations, such as grouping the nodes in the \ac{sg3d}. Third, the performance of our system is limited to the 3D reconstruction and segmentation methods to construct \ac{sg3d}, and it currently operates at the object level. Future efforts could focus on expanding its capabilities to encompass more levels of perception and understanding. Finally, while \modelname's zero-shot reasoning capability can generalize to novel scenes, its improved representation after alignment only works in one specific scene. Leveraging its interaction with humans to enhance its adaptability and scalability to novel scenes is one important future direction.

\section{Conclusion}
\label{sec:conclusion}
In this paper, we present \modelname, a novel framework designed to achieve human-robot collaboration and bridge their perceptual gap. \modelname leverages 3D reconstruction to create \acl{sg3d} as innate representations, and decomposes complex tasks and solve them sequentially through natural interactions with humans. Experimental results validate \modelname's capability in zero-shot 3D reasoning, achieving competitive performance on the ScanQA benchmark without alignment. Furthermore, our alignment experiments highlight its proficiency in achieving human-robot alignment across varying levels of task difficulty with high user satisfaction ratio and transferability to novel tasks. We hope our efforts and insights could facilitate the deployment of LLM-based robot systems in real-world scenarios.

\section*{Acknowledgement}
The authors would like to thank Huangyue Yu (BIGAI) for her help on 3D scene graph generation, and Ziyu Zhu (Tsinghua University, BIGAI) for his help on 3D semantic segmentation. The authors would also like to thank the subjects for their efforts during the alignment human study, and other colleagues from BIGAI fruitful discussions.

\bibliographystyle{IEEEtran} 
\bibliography{reference}

\end{document}

%% file: config.tex
\usepackage{acronym}
\usepackage{xspace}
\usepackage{graphicx}
\usepackage{caption}
\usepackage{amsmath}
\usepackage{xcolor}
\usepackage{subcaption}
\usepackage{booktabs}
\usepackage{comment}
\usepackage{tabularx}
\usepackage{multirow}
\definecolor{LightGray}{gray}{0.9}
\definecolor{cvprblue}{rgb}{0.21,0.49,0.74}
\usepackage[pagebackref,breaklinks,colorlinks=true,linkcolor=cvprblue,citecolor=cvprblue,bookmarks=false]{hyperref}
\usepackage[capitalise,nameinlink]{cleveref}
\usepackage[sort]{cite}

\newcommand{\etc}{\emph{etc.}\xspace}
\newcommand{\eg}{\emph{e.g.}\xspace}
\newcommand{\ie}{\emph{i.e.}\xspace}

\newcommand{\sota}{\text{state-of-the-art}\xspace}

\acrodef{llm}[LLM]{large language model}
\acrodef{mmlm}[MMLM]{multi-modal language model}
\acrodef{sft}[SFT]{supervised fine-tuning}
\acrodef{rlhf}[RLHF]{Reinforcement Learning from Human Feedback}
\acrodef{sg3d}[3DSG]{3D Scene Graph}
\acrodef{qa}[QA]{question-answering}
\acrodef{gui}[GUI]{graphical user interface}
\acrodef{cot}[CoT]{Chain-of-Thought}

\newcommand{\modelname}{\textsc{SynergAI}\xspace}
\newcommand{\srh}{SR$_{\text{Human}}$}
\newcommand{\srl}{SR$_{\text{LLM}}$}
\newcommand{\sri}{SR$_{\text{Init}}$}

\newcommand{\rrr}{RR$_{\text{Interaction}}$}

\definecolor{gblue}{HTML}{4285F4}
\definecolor{gred}{HTML}{DB4437}
\definecolor{ggreen}{HTML}{0F9D58}
\definecolor{vblue}{HTML}{2993ba}

%% file: math_commands.tex
\usepackage{amsmath,amsfonts,bm}

\def\eqref#1{equation~\ref{#1}}

\def\1{\bm{1}}

\DeclareMathAlphabet{\mathsfit}{\encodingdefault}{\sfdefault}{m}{sl}
\SetMathAlphabet{\mathsfit}{bold}{\encodingdefault}{\sfdefault}{bx}{n}



%% file: authors.tex
\author{Yixin Chen$^{*}$, Guoxi Zhang$^{*}$, Yaowei Zhang$^{*}$, Hongming Xu$^{*}$, Peiyuan Zhi, Qing Li$^\dagger$, Siyuan Huang$^\dagger$
\\ \textbf{\url{https://synerg-ai.github.io/}}%
\thanks{$^{*}$ Equal Contribution.}
\thanks{$^\dagger$ Corresponding Authors.}
\thanks{State Key Laboratory of General Artificial Intelligence, Beijing Institute for General Artificial Intelligence (BIGAI)}%
}
